\documentclass[journal]{IEEEtran}

\newif\ifproofs
\usepackage{mathtools, amsthm, amsmath,amssymb,amsfonts, dsfont, cancel}
\usepackage{textcomp, acronym, cite, url, comment, footnote, dirtytalk}
\usepackage{graphicx, xcolor, enumitem, xspace, diagbox, adjustbox ,fancyhdr, multicol, multirow}
\usepackage[caption=false,font=normalsize,labelfont=sf,textfont=sf]{subfig}
\usepackage[bookmarks,colorlinks]{hyperref} % pagebackref=true,
\usepackage[linesnumbered,ruled,vlined]{algorithm2e}
\let\oldnl\nl% Store \nl in \oldnl
\newcommand{\nonl}{\renewcommand{\nl}{\let\nl\oldnl}} % Remove line
\setlist[description]{font=\normalfont}
\def\BibTeX{{\rm B\kern-.05em{\sc i\kern-.025em b}\kern-.08em %\rm definition
T\kern-.1667em\lower.7ex\hbox{E}\kern-.125emX}}
\usepackage[noend]{algpseudocode}
\usepackage{accents}

\acrodef{ml}[ML]{machine learning}
\acrodef{fl}[FL]{federated learning}
\acrodef{sgd}[SGD]{stochastic gradient descent}
\acrodef{nn}[NN]{neural network}
\acrodef{fa}[FedAvg]{federated averaging} 
\acrodef{snr}[SNR]{signal-to-noise ratio}
\acrodef{mlp}[MLP]{multi-layer perceptron}
\acrodef{dnn}[DNN]{deep neural network}
\acrodef{cnn}[CNN]{convolutional neural network}
\acrodef{salf}[SALF]{straggler-aware layer-wise federated learning}

\newcommand{\user}{u}
\newcommand{\users}{U}
\newcommand{\wOpt}{\vw_{\rm opt}}
\newcommand{\sample}{i^u_t}
\newcommand{\ltUsers}{\cU_t^l}
\newcommand{\depth}{d_{t}^{u}}
\newcommand{\cD}{\mathcal{D}}
\newcommand{\cU}{\mathcal{U}}

\newcommand{\cO}{\mathcal{O}}

\newcommand{\R}{\mathbb{R}}

\newcommand{\E}{\mathds{E}}
\newcommand{\prob}{\mathds{P}}

\newcommand{\vw}{\boldsymbol{w}}

\newcommand{\vx}{\boldsymbol{x}}

\DeclareMathOperator*{\Bin}{{\rm Bin}}

\DeclarePairedDelimiter\abs{\lvert}{\rvert}

\DeclareMathOperator*{\argmin}{arg\,min}

\newtheorem{theorem}{Theorem}[section]

\newtheorem{lemma}{Lemma}

\title{Stragglers-Aware Low-Latency Synchronous Federated Learning via Layer-Wise Model Updates}

\author{\IEEEauthorblockN{Natalie Lang, Alejandro Cohen, and Nir Shlezinger\\} 
\thanks{This work was partially supported by the Israeli Ministry of Science and Technology. 
N. Lang and N. Shlezinger are with the School of ECE, Ben-Gurion University of the Negev,  Be’er-Sheva, Israel (e-mails:  langn@post.bgu.ac.il; nirshl@bgu.ac.il). A. Cohen  is with the Faculty of ECE, Technion – Israel Institute of Technology, Haifa, Israel (e-mail: alecohen@technion.ac.il).}  		
}

\begin{document}
\maketitle
\begin{abstract}
Synchronous \ac{fl} is a popular paradigm for collaborative edge learning. It typically involves a set of heterogeneous devices locally training \ac{nn} models in parallel with periodic centralized aggregations. As some of the devices may have limited computational resources and varying availability, \ac{fl} latency is highly sensitive to stragglers. Conventional approaches  discard incomplete intra-model updates done by stragglers, alter the amount of local workload and architecture, or resort to asynchronous settings; which all affect the trained model performance 
% of the trained model when operating 
under tight training latency constraints.
In this work, we propose {\em \ac{salf}} that leverages the optimization procedure of \acp{nn} via backpropagation to update the global model in a {\em layer-wise} fashion. \ac{salf} allows stragglers to synchronously convey partial gradients, having each layer of the global model be updated independently with a different contributing set of users. We provide a theoretical analysis, establishing convergence guarantees for the global model under mild assumptions on the distribution of the participating devices,  revealing that \ac{salf}
%Our analysis reveals that \ac{salf}-aided \ac{fl} of \acp{nn} is proven to 
converges at the same asymptotic rate as
\ac{fl} with no timing limitations. This insight is matched with empirical observations, demonstrating the performance gains of \ac{salf} compared to alternative mechanisms mitigating the device heterogeneity gap in \ac{fl}.
%, and its ability to facilitate synchronous \ac{fl} with low latency constraints without notably affecting the utility of the learned model.
\end{abstract}
\acresetall

\section{Introduction}\label{sec:intro}
% importance of FL
\IEEEPARstart{D}{eep} learning algorithms require large volumes of data. In practice, data is often gathered by edge devices such as mobile phones, sensors, and vehicles, which may be limited in their ability to share this data, due to, e.g., privacy or regulation constraints \cite{horvitz2015data}. 
{\em\Ac{fl}}~\cite{mcmahan2017communication,kairouz2021advances, li2020federated, gafni2021federated} is an emerging paradigm that allows multiple devices to learn a model collaboratively. To avoid data sharing, \ac{fl} exploits the local computational capabilities of edge users \cite{chen2019deep}, training locally with periodic aggregations orchestrated by a server. 

% challengaes of FL
Prevalent \ac{fl} protocols are synchronous, operating in multiple rounds. In each round, the server transmits the latest update of the global model to all participating clients. Each client then locally trains the model using its local compute power and data set, and subsequently sends the updated version back to the server for aggregation. 
This distributed operation of \ac{fl} inherently induces several core challenges that are not encountered in conventional centralized learning \cite{gafni2021federated,li2020federated}. A notable challenge is associated with the {\em latency} of learning in a federated manner, which is a dominant factor, particularly in \ac{fl} applications that operate continuously in dynamic environments and must thus learn rapidly. Such applications include, e.g., intelligent transportation systems~\cite{samarakoon2019distributed}, and wireless networks adaptation \cite{xia2021federated, lu2020low, shi2020joint}. 
 
The excessive latency of \ac{fl} is mainly due to two main factors: $(i)$ the time it takes to communicate the model updates between the users and the server; and $(ii)$ the local computation time at the users' side. The former is typically addressed by different forms of sparsification~\cite{han2020adaptive,hardy2017distributed, aji2017sparse, alistarh2018convergence} and compression ~\cite{shlezinger2020uveqfed, alistarh2017qsgd,reisizadeh2020fedpaq, lang2023compressed}. Thus, the dominant factor is often the latter, i.e., the time it takes each user to update the local model. This issue is most significant in \ac{fl} settings operating under {\em system heterogeneity}, arising from the existence of devices with low computational capabilities~\cite{diao2020heterofl,pfeiffer2023federated}, which may even be temporally unavailable while the local training takes place \cite{vahidian2023curricula}.  
The heterogeneous nature of edge devices induces possibly substantial variations between different clients in their local update latency. This in turn affects the time it takes the server to update the global model on each round. As a result, \ac{fl} is sensitive to {\em stragglers}, as each round takes as long as the local training time of the slowest user, dictating latency and throughput implications \cite{kairouz2021advances}. The presence of heterogeneous users thus makes synchronous \ac{fl} abortive for applications with tight latency constraints. 

% existing methods
Various schemes were proposed to provide robustness against stragglers in \ac{fl}~\cite{nishio2019client,reisizadeh2022straggler,schlegel2023codedpaddedfl,wang2023fluid,li2020heterogenous,park2021sageflow,chen2020asynchronous}, see also survey~\cite{pfeiffer2023federated}. Conventional synchronous \ac{fl} limits local computation latency by imposing a deadline, while discarding a fixed-size set of delayed stragglers and their contributions~\cite{bonawitz2019towards, li2019convergence}. 
This can be extended to support varying deadlines via user selection, where only a subset of the users participate in each round~\cite{chen2021communication}, by identifying and grouping potential stragglers~\cite{nishio2019client,reisizadeh2022straggler}. Deadline-based synchronous \ac{fl} facilitates the incorporation of latency constraints  without altering the learning  procedure, and can be combined with additional \ac{fl} latency reduction techniques based on scheduling~\cite{xia2021federated} and resource allocation~\cite{shi2020joint}. However, the fact that stragglers are discarded affects performance when operating under low latency requirements, where a large portion of the users may not meet the local computation deadline.
 
Alternative approaches to handle stragglers involve deviating from the conventional operation of \ac{fl} by either altering the learning procedure, or by switching to an asynchronous operation \cite{pfeiffer2023federated}. Several examples for the former are dedicated aggregation \cite{li2020heterogenous}; introducing redundancy on the devices’ data via distributed codes~\cite{esfahanizadeh2022stream,schlegel2023codedpaddedfl}; and altering the amount of local workload and architecture \cite{diao2020heterofl, wang2023fluid}. 
However, these approaches assume a limited portion of straggling users and are thus unsuitable for low-latency settings, where many users might struggle to locally compute sufficient local training iterations in time. 
% asynchronous
The asynchronous approach operates \ac{fl} without requiring participating users to aggregate on each round~\cite{park2021sageflow,chen2020asynchronous, nguyen2022federated}. 
For instance,  TimelyFL  proposed in \cite{zhang2023timelyfl} was experimentally shown to mitigate stragglers in asynchronous settings by having the server and the clients share calculation times to dictate a latency and local workload while utilizing layer-wise aggregations.
Whereas asynchronous \ac{fl}  allows slower clients to continue the local training and contribute to future aggregation rounds, it  also requires the number of slow computing users to be small for stable learning~\cite{pfeiffer2023federated}, limiting its applicability under tight latency constraints. Moreover, having different nodes operate on different versions of the global model leads to staleness \cite{ortega2023asynchronous}, complicating the orchestration of the \ac{fl} procedure compared to synchronous \ac{fl}, while often ending up with an inferior performance.   

% in this work
% high-level
In this work, we propose {\em \ac{salf}}, for \ac{fl} with time varying system heterogeneity, that enables synchronous deadline-based high-performance low-latency operation. \ac{salf} is particularly geared for learning \acp{dnn}, being the common family of machine learning models considered in \ac{fl}, while exploiting their gradient-based training operation. 
We leverage the inherent recursive nature in which \ac{dnn} are being optimized, i.e., the fact that the empirical risk gradient with respect to the model weights is computed from the last layer to the first via backpropagation~\cite{rumelhart1985learning}, and that gradient computation typically induces a dominant portion of the latency. This operation indicates that stragglers may still compute last-layers gradients that can be utilized rather than discarded under strict deadlines. \ac{salf} is thus based on this layer-wise approach, allowing users to convey  (possibly partial) gradients when local training expires. Then, to update the global model, \ac{salf} averages the local updates {\em per layer}, while preserving the simplicity of conventional \ac{fa} \cite{mcmahan2017communication}, and having different contributing devices for each.

%in detail
To capture the realistic dynamic heterogeneity of the users, we analyze \ac{salf} assuming a probabilistic model on the straggling clients IDs per round. Specifically, in each round, the stragglers' set ranges from being empty to contain all \ac{fl} users, neither limiting its cardinality nor its items. In our analysis we rigorously prove that \ac{salf} converges to the optimal model in the same asymptotic rate as local \ac{sgd}~\cite{li2019convergence}, while characterizing an upper bound on the gap in its learning objective value compared to the optimal model in the non-asymptotic regime. 

We extensively evaluate \ac{salf} for the federated training of different \ac{dnn} architectures, considering various latency constraints.  Our numerical results show that \ac{salf} allows reliable training under tight latency constraints where a large bulk of the users become stragglers, while achieving similar accuracy to conventional \ac{fa} with no latency requirements. 

% organization
The rest of this paper is organized as follows: Section~\ref{sec:system_model} briefly reviews the \ac{fl} system and the heterogeneity models. Section~\ref{sec:method} presents \ac{salf} along with its convergence analysis.  We numerically evaluate \ac{salf} in Section~\ref{sec:experiments}, and provide concluding remarks in Section~\ref{sec:conclusions}.

% Notations....
Throughout this paper, we use boldface lower-case letters for vectors, e.g., $\vx$, and calligraphic letters for sets, e.g., $\mathcal{X}$, with $\abs{\mathcal{X}}$ being the cardinality of $\mathcal{X}$. The stochastic expectation, probability operator, and $\ell_2$ norms are denoted by $\E[\cdot]$, $\prob[\cdot]$,  and $\|\cdot\|$, respectively, while  $\R$ is the set of  real numbers.

\section{System Model}\label{sec:system_model}
In this section, we set the ground for the derivation of \ac{salf}. We commence by presenting the system model of synchronous \ac{fl} in Subsection~\ref{subsec:fl}. Then,  we provide a description of the system heterogeneity model and its effects on local computation latency in Subsection~\ref{subsec:system_heterogeneity_model}.

\begin{figure*}
\centering    
    \includegraphics[width=\textwidth]{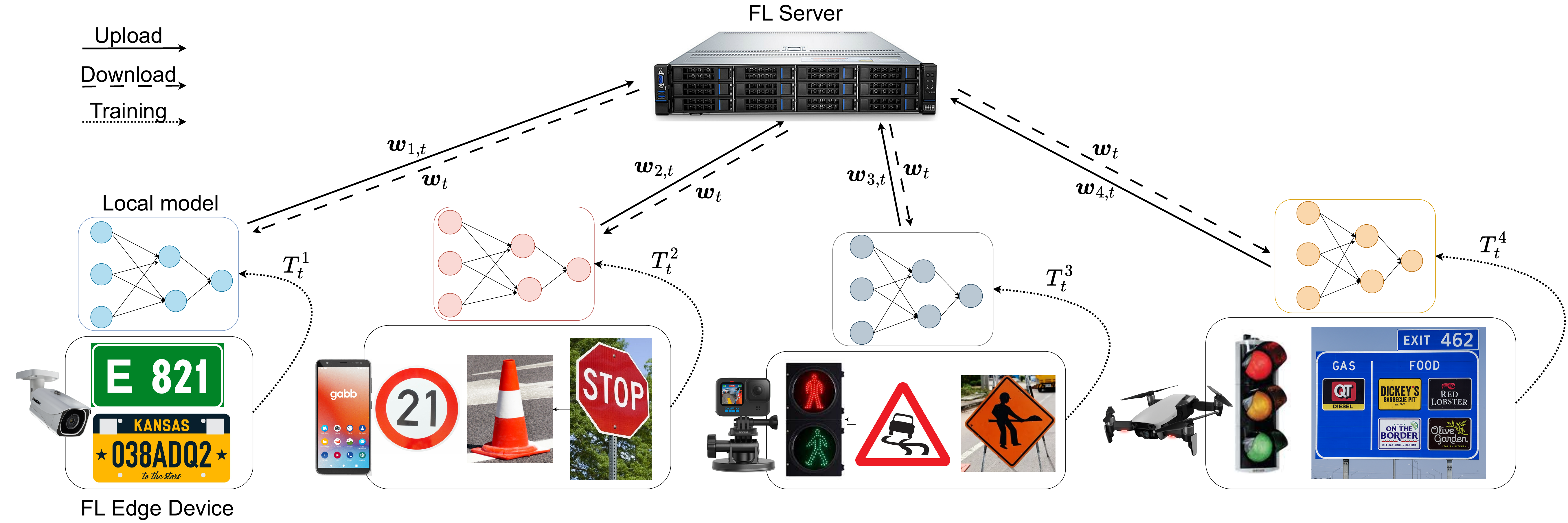}
    \caption{A device-heterogeneous \ac{fl}-aided system learning object recognition that is expected to operate under tight latency and edge power constraints. Note that $\vw_t$ and $\vw_{\user,t}$ denote the global and local models, respectively; where $T^\user_t$ it the local computational time of user $\user$ in \ac{fl} round $t$.}   
    \label{fig:FL_ObjRec}   
\end{figure*}

\subsection{Federated Learning}\label{subsec:fl}
% FL definition
We consider a central server training a model with parameters $\vw\in \R^m$ using data available at $\users$ users, where each is indexed by $ \user \in \{1,\dots, \users\}$. Unlike conventional centralized learning, these datasets, denoted $\cD_1,\dots, \cD_\users$, cannot be shared with the server.  
% mathematical formulation
Thus, by letting $F_\user(\vw)$ be the empirical risk of a model $\vw$ evaluated with dataset $\cD_\user$,
% , i.e.,% \begin{align}\label{eq:local_objective_def}
%     F_k(\vw) = \frac{1}{|\cD_k|} \sum_{(\vx_i,y_i) \in \cD_k} \ell(\vx_i,y_i;\vw),     
% \end{align}
% where it is assumed in \eqref{eq:local_objective_def} that $\cD_k$ is labeled and $\ell(\cdot)$ stands for the user-specified loss function.
\ac{fl} aims to recover the $m\times 1$ optimal weights vector, $\wOpt$, satisfying
\begin{equation}\label{eq:w_opt}
    \wOpt = \argmin_{\vw} \left\{F(\vw)\triangleq \sum_{\user=1}^\users \frac{1}{\users} F_\user\left(\vw\right)\right\},
\end{equation}
% FL steps
where it is implicitly assumed that the local datasets are of balanced (equal) cardinality.

Generally speaking, \ac{fl} operates in rounds where for each time step $t$, the server distributes the global model $\vw_t$ to the users, who each locally trains it, and sends back the model updates~\cite{gafni2021federated}.  %The users thus do not directly expose their private data as training is performed locally. 
In conventional synchronous \ac{fl}, the server collects the model updates from all the participating users,  aggregates the models into an updated global model, and the overall procedure repeats iteratively. 

% local optimization
We focus on settings where $\vw$ represents a \ac{dnn} with gradient-based local training. Here, each user of index $\user$ computes the stochastic gradient of its local empirical risk $F_\user$ evaluated on the global model at time $t\geq 1$, $\vw_t$; i.e., $\nabla F_\user\left(\vw_t;\sample\right)$, where $\sample$ denotes the data sample index, chosen uniformly from $\cD_\user$. 
Then, for a step-size $\eta_t$, the user shares its local update (gradients), i.e,
\begin{align}\label{eq:user_update}
    \vw_{\user,t} \triangleq \vw_t - \eta_t\nabla F_\user\left(\vw_t;\sample\right),
\end{align}
with the server who updates the global model. 
%
%aggregation 
Conventional aggregation of the local updates is based on \ac{fa}~\cite{mcmahan2017communication}, in which the server sets the global model to be 
\begin{align}\label{eq:FedAvg_update}
    \vw_{t+1} \triangleq \sum_{\user=1}^\users \frac{1}{\users}\vw_{\user,t} 
            = \vw_t - \eta_t\sum_{\user=1}^\users\frac{1}{\users}\nabla F_\user\left(\vw_t;\sample\right).
\end{align} The updated global model is again distributed to the users, and the learning procedure continues until convergence is reached. 

\ac{fl} of \acp{dnn} involves edge users training, where each user obtains its local gradients in \eqref{eq:FedAvg_update} using its local computational capabilities. The  users' devices can notably vary  in their computational resources based on their hardware and instantaneous operation. This property gives rise to the core challenge of device heterogeneity gap, discussed next.

\subsection{System Heterogeneity}\label{subsec:system_heterogeneity_model}
Edge devices participating in \ac{fl} can widely differ in their computational powers, leading to varying processing times for calculating gradients
\cite{reisizadeh2022straggler}. Specifically, let $T_t^\user$ denote the time it takes the $\user$th user to compute the gradient at the $t$th round, i.e., $\nabla F_\user\left(\vw_t ;\sample\right)$. Higher values of $T_t^\user$ are attributed with clients who are slower at the $t$th round due to, e.g., limited hardware or additional external computations being carried out. These slower users are termed {\em stragglers}. 
Hence, the local computation latency of the $t$th \ac{fl} round is given by $ \max_{\user}
T_t^\user$, as the server has to {wait} for the slowest user. This limits \ac{fl} applications with tight latency constraints, e.g., intelligent transportation systems~\cite{zhang2023federated}, as illustrated  in Fig.~\ref{fig:FL_ObjRec}. 
%which illustrates a device-
% heterogeneous \ac{fl}-aided intelligent transpiration system for the task of object recognition. There, the high dynamic nature of vehicles places extremely high real-time requirements for their decision-making, and excessive response latency might seriously jeopardise safety \cite{zhang2023federated}. A representative scenario for an excessive response latency is a server waiting for a group of power-constrained devices, having short-time availability before being turned-off. 

A requirement for executing an \ac{fl} round with a fixed latency is satisfied by setting a deadline $T_{\max}$. Deadline-based synchronous \ac{fl} typically discards the stragglers at round $t$ for which $T_t^\user > T_{\max}$, i.e., those not meeting the threshold~\cite{bonawitz2019towards}. In such cases, 
denoting the set of users meeting the deadline at round $t$ as $\mathcal{U}_t \triangleq \{\user: T_t^\user \leq T_{\max}\}$,
the aggregation rule in \eqref{eq:FedAvg_update} is replaced with
\begin{align}\label{eq:fl_updateStr}
\vw_{t+1} =\vw_t - \eta_t \sum_{u \in \mathcal{U}_t} \frac{1}{\abs{\mathcal{U}_t}}\nabla F_\user\left(\vw_t;\sample\right).
\end{align}
Aggregation via~\eqref{eq:fl_updateStr} guarantees that each round does not surpass the desired deadline. However, for low-latency, i.e., small $T_{\max}$, this could result in a few users participating in each round, degrading the training procedure. Such a scenario motivates the deriviation of a scheme which allows \ac{fl} to operate with small $T_{\max}$ without fully discarding stragglers, as proposed next.

% diagram block figure
\begin{figure*}
     \centering             
     \includegraphics[width=\textwidth]{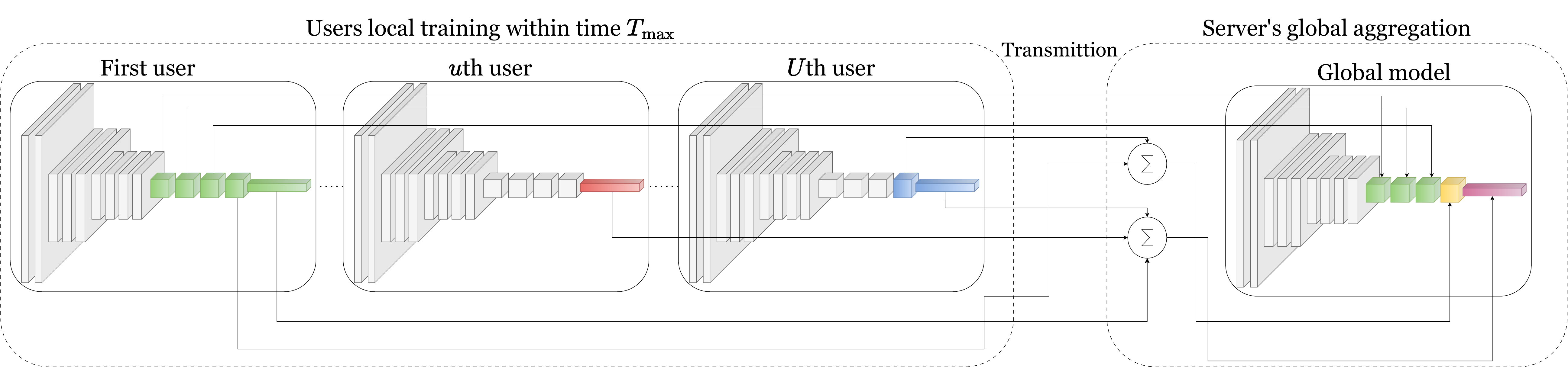}
    \caption{Illustrative overview of \ac{salf} for training a deep \ac{cnn}. The left dashed-box represents local training, where colored layers correspond to gradients calculated within $T_{\max}$;  the left dashed-box shows the layer-wise aggregation with updated colored global model layers.}\label{fig:diagram_block}
\end{figure*}

\section{Straggler-Aware Layer-Wise FL}\label{sec:method}
In this section, we introduce \ac{salf}, formulating its operation in Subsection~\ref{subsec:algorithm}. Then, in Subsection~\ref{subsec:analysis}  we analyze the convergence properties of \acp{dnn} trained in a federated manner using \ac{salf}, and provide a discussion in Subsection~\ref{subsec:discussion}. For clarity, we summarize the symbols and notations used throughout this section in Table~\ref{tbl:notations}.

% motivation
\subsection{SALF Algorithm}\label{subsec:algorithm}
Consider the gradient-based federated training of an $L$-layered \ac{dnn} with layer-wise parameters  
\begin{align}\label{eq:global_layers}
\vw_t = \left[\vw^1_t, \dots, \vw^L_t\right]^T,    
\end{align}
where layers $1$ and $L$ correspond to the input and output layers, respectively. 
The local gradients $\nabla F_\user$ are typically computed using backpropagation \cite{rumelhart1985learning}. Namely, gradients are recursively calculated from the last-to-first layer, constructing
\begin{equation}
    \big[\nabla F^{1}_\user(\vw_t;\sample), \dots,\nabla F^l_\user(\vw_t;\sample),\dots \nabla F^L_\user(\vw_t;\sample)\big]^T, 
\end{equation}
each component at a time, where $F^l_\user(\vw_t;\sample)$ denotes the (stochastic) gradient computed with respect to the parameters of the $l$th layer.
This operation suggests that when $T_{\max}$ expires, stragglers are likely to evaluate partial gradients, corresponding to the last \ac{dnn} layers. 

To exploit this expected behavior, we design \ac{salf} {\em not to discard stragglers}, but to have them {\em contribute to the aggregation of these intermediate layers}. This is done by aggregating via \ac{fa} in \eqref{eq:FedAvg_update} over each layer separately. 
Since a device's computational power availability for an \ac{fl} round is assumed to diverse along the training, the depth reached in backpropagation when the deadline expires is expected to vary between users and rounds. In our analysis provided in the sequel, it is modeled as  a discrete random variable obeying a uniform distribution.

\ac{salf}, illustrated in Fig.~\ref{fig:diagram_block} and summarized as Algorithm~\ref{alg:salf}, affects two aspects of conventional synchronous \ac{fl}: $(i)$ {\em local training} at the users side; and $(ii)$ {\em aggregation} of the model updates by the server.  The resulting operation is  formalized below.

% Algorithm summary
\SetKw{Initialization}{Initialization:}
\SetKwBlock{Users}{Users side:}{end}
\SetKwBlock{Server}{Server side:}{end}
\SetKwBlock{DoParallel}{do in parallel for each $\user$, until deadline $T_{\max}$:}{end}
\begin{center}
\begin{algorithm}
\caption{\ac{salf} at round $t$}
\label{alg:salf}
 \Initialization{\text{FL round running time $T_{\max}$}\;}
    \Users{
    \DoParallel{
           Compute $\nabla F_\user\left(\vw_t,\sample\right)$ up to layer $\depth$;\\
           Convey partial gradients to server; 
    }}
    \Server{
    \For{$1\leq l \leq$ L}{
            Recover $\cU_t^l$; \Comment{the users updating the $l$th layer}\\
           Compute $\Tilde\vw^l_{t+1}$ via \eqref{eq:salf_update_rule};
           \Comment{layer-wise update}
           }}
   \KwResult{The updated global model, $\Tilde\vw_{t+1}$;}
\end{algorithm}
\end{center}
 
\subsubsection{Users Local Training}
As in conventional \ac{dnn} training, each user computes $\nabla F_\user\left(\vw_t;\sample\right)$ via backpropagation, sequentially, from the last layer to the first. When the deadline $T_{\max}$ expires, the $\user$th user calculates the gradients up to layer $\depth \in \{1,\dots,L,L+1\}$, denoting its associated depth; where $\depth=L+1$ stands for a straggler that does not compute any layer gradient. 

Accordingly, if $\depth\leq L$, the $\user$th user conveys to the server the $L-\depth+1$ sub-vectors of its empirical risk stochastic gradient evaluated on $\vw_t$, i.e.,
\begin{equation}\label{eqn:PartGrad}
    \big[\nabla F^{\depth}_\user(\vw_t;\sample), \dots, \nabla F^L_\user(\vw_t;\sample)\big]^T. 
\end{equation}
This operation implies that users that are typically viewed as stragglers also convey (partial) model updates to the server.

\smallskip
\subsubsection{Server Aggregation}
Let $\ltUsers \subseteq \{1,\ldots,\users\}$ be the set of users that managed to compute the $l$th layer gradient on round $t$. Using the above notations, this set is given by 
\begin{align}\label{eq:ltUsers}
\ltUsers \triangleq \{\user: \depth\leq l\},    
\end{align}
where $\mathcal{U}^1_t\subseteq \dots \subseteq \ltUsers \subseteq \dots \subseteq \mathcal{U}^L_t$. This follows by the backpropagation operation, as if a gradient was calculated up to layer $\depth<L$, so do all the gradients of layers $\depth+1,\dots,L$.

The server recovers ${\{\ltUsers\}}_{l=1}^L$ from the received gradients, which are then aggregated via a form of {\em layer-wise} \ac{fa}. 
Rather than using \ac{fa} over the full \ac{dnn} as in~\eqref{eq:FedAvg_update}, the model updates obtained by \ac{salf}, denoted by $\Tilde\vw_{t+1}$, are aggregated in a layer-wise fashion. Specifically, the $l$th layer of the global model, according to the decomposition in~\eqref{eq:global_layers}, is updated via  
\begin{align}\label{eq:salf_update_rule} 
\Tilde{\vw}^l_{t+1}=
\begin{cases}
\vw^l_t & \abs{\ltUsers}=0, \\
\frac{1}{1-p_l}\left(\sum_{\user\in \ltUsers} \frac{1}{\abs{\ltUsers}}\vw^l_{\user,t} 
-p_l\vw^l_t\right) & \abs{\ltUsers}>0; \\
\end{cases}
\end{align}
where $p_l\in[0,1)$ is a hyperparameter set such that $\Tilde{\vw}^l_{t+1}$ is an unbiased estimator of the corresponding stragglers-free \ac{fa} update $\vw^l_{t+1}$. For instance, when the $\{\depth\}$ are i.i.d with the uniform distribution, $p_l$ is given by 
\begin{align}\label{eq:p_constant}   
p_l= {\left(1-\frac{l}{L+1}\right)}^\users,
\end{align}
see further details in Subsection~\ref{subsec:analysis}.

Note that in \eqref{eq:salf_update_rule}, as opposed to conventional \ac{fa} \eqref{eq:FedAvg_update}, the set of participating users can vary between layers. This is because, when operating under a deadline that imposes a fixed latency, different users compute their gradients up to different layers (see Fig.~\ref{fig:diagram_block}), depending on their computational resources and availability at that round, encapsulated in the stochastic $\depth$. As each layer of the global model is updated by different users at different rounds, the convergence analysis of \ac{salf} deviates from traditional \ac{fl} \cite{li2019convergence}. Yet, due to the formulation of the model updates in \eqref{eq:salf_update_rule}, we are still able to rigorously prove convergence under conventional modelling assumptions, as reviewed next.

% notations summary
\begin{table}
\caption{Summery of notations}
\label{tbl:notations}
\centering
%\begin{adjustbox}{width=\columnwidth} 
\begin{tabular}{c|c}
\hline
Symbol & Description \\
\hline
$\users, \user$ & index of the clients\\
$L, l$ & index of \ac{dnn} layer\\
$t$ & \ac{fl} global iteration (round) index\\
$\eta_t$ & learning rate at time $t$\\
$T_{\max}$ & latency constraint per round\\
$\ltUsers$ & set of clients indices updating the $l$th layer at time $t$\\
$\sample$ & data sample index of the $\user$th client at time $t$\\
$\nabla F_\user\left(\vw_t;\sample\right)$ & stochastic gradient of the $\user$th client at time $t$  \\
$\depth$ & depth reached in backpropagation by client $\user$ at time $t$ \\
$\vw_t, \vw^l_t$ & \ac{fa} full and $l$th sub vectors at time $t$\\
$\vw_{t,\user}$ & $\user$th client
local model at time $t$ \\
$\Tilde\vw_t$ & \ac{salf} global model at time $t$\\
$p_l$ & a hyperparameter assuring the unbiaesness of \ac{salf}\\
$\rho_s, \rho_c$ & smoothness and convexity constants\\
$\sigma^2_\user, G^2$ & stochastic gradient related constants\\
\hline
\end{tabular}
%\end{adjustbox}
\end{table}

\subsection{Analysis}\label{subsec:analysis}
As described in the previous subsection, \ac{salf} tackles system heterogeneity using a layer-wise aggregation rule to update the global model in each iteration. Here, we theoretically characterize the convergence profile of \ac{salf}. Note that the layer-wise operation of \ac{salf} and the non-zero probability of having the first layers not updated in given rounds indicate that existing \ac{fl} analysis with partial device participation, e.g., \cite[Thm. 3]{li2019convergence}, do not apply here.
%exploring \ac{fl} convergence for partial device participation. Whereas \cite{li2019convergence} performs aggregation over one random set of users, \ac{salf} would have multiple of those, as the number of layers of the trained \ac{dnn}. 
%
We first elaborate bellow the assumptions introduced in our analysis as well as the chosen statistical characteristics of the stragglers; from which the convergence bound is subsequently derived.

\smallskip
\subsubsection{Analysis Assumptions} We carry out our analysis of \ac{salf} subject to the following assumptions, that are commonly employed in  \ac{fl} convergence studies \cite{stich2018local, shlezinger2020uveqfed, lang2023joint, lang2023compressed}:
\begin{enumerate}[label={\em AS\arabic*},series=assumptions]       
    \item \label{itm:smooth_strng_convx_lip_cont} 
    The local objectives $\{F_\user(\cdot)\}^\users_{\user=1}$ are  $\rho_c$ strongly convex and $\rho_s$-smooth. That is, 
    for all $\vw_1, \vw_2 \in \R^m$, it holds that
    \begin{align*}       
     (\vw_1 -\vw_2)^T\nabla F_\user(\vw_2)&+\frac{1}{2}\rho_c{\|\vw_1 -\vw_2\|}^2 \\
      \leq    F_\user(\vw_1)&-F_\user(\vw_2)  \leq \\
        (\vw_1 -\vw_2)^T\nabla F_\user(\vw_2)&+\frac{1}{2}\rho_s{\|\vw_1 -\vw_2\|}^2.
    \end{align*}
    \item \label{itm:bounded_var}  For each user $u$ and round index $t$, the variance of the stochastic gradients $\nabla F_\user(\vw;\sample)$ is bounded by some $\sigma^2_\user$ for all $\vw\in\R^m$, i.e.,
    $$ \E\left[{\|\nabla F_\user(\vw;\sample) - \nabla F_\user(\vw)\|}^2\right]\leq \sigma^2_\user.$$
    \item \label{itm:bounded_norm} For each user $u$ and round index $t$, the expected squared $\ell_2$ norm of the stochastic gradients $\nabla F_\user(\vw;\sample)$ is uniformly bounded by some $G^2$ for all $\vw\in\R^m$, i.e.,
    $$ \E\left[{\|\nabla F_\user(\vw;\sample)\|}^2\right]\leq G^2.$$
    \item \label{itm:random_stragglers} For each user $u$ and round index $t$, the depth reached in backpropagation, denoted $\depth,$ is i.i.d. (in $t$ and $u$), and uniformly distributed over the set $\{1,\dots,L+1\}$.
\end{enumerate}

\smallskip
Notice that smoothness, as assumed in \ref{itm:smooth_strng_convx_lip_cont}, holds for a range of objective functions used in \ac{fl}, including $\ell_2$-norm regularized linear/logistic regression \cite{shlezinger2020uveqfed}. 
The heterogeneity between the users is also reflected in \ref{itm:bounded_var} via the dependence on $\user$, which implies that the local objectives differ between users, as different datasets can be statistically heterogeneous, i.e., arise from different distributions.
Accordingly, we follow \cite{li2019convergence} and define the (datasets) heterogeneity gap as
\begin{align}\label{eq:psi_heterogeneity_gap}
\Gamma \triangleq F(\wOpt)-\frac{1}{\users}\sum_{\user=1}^\users \min_{\vw} F_\user(\vw),
\end{align}
where $\wOpt$ is given in~\eqref{eq:w_opt}. 

\smallskip
\subsubsection{Stragglers Statistical Modelling}
The dynamic nature of the system's devices heterogeneity (see Subsection~\ref{subsec:system_heterogeneity_model}) is statistically modelled in~\ref{itm:random_stragglers}, as the set of random variables $\{\depth\}$ in~\ref{itm:random_stragglers}  form the stragglers sets $\{\ltUsers\}$ \eqref{eq:ltUsers}. This formulation subsumes the extreme cases of both $\abs{\ltUsers}=0$ (a layer which is not updated by any user) and $\abs{\ltUsers}=\users$ (a layer that is updated by all users). 
In particular, the marginal distribution of $\abs{\ltUsers}$ is obtained in the following lemma:

%the i.i.d. uniform distribution of $\{\depth\}$  in fact induces a precise distribution on $\abs{\ltUsers}$, stated as the following observation.

\begin{lemma}\label{obs:binom_rv}
When \ref{itm:random_stragglers} holds, then for every \ac{fl} round $t$ and \ac{dnn} layer $l$, and the cardinality of the set $\ltUsers$ is distributed as
\begin{align}\label{eq:binom_rv}
    \abs{\ltUsers}\sim \Bin\left(\users,\frac{l}{L+1}\right),
\end{align}
where $\Bin$ denotes the Binomial distribution.
\end{lemma}
\begin{IEEEproof}
Eq.~\eqref{eq:binom_rv} follows from the definition of  $\ltUsers$ in \eqref{eq:ltUsers}, which implies that for every $M \in \{0,\ldots,\users\}$, it holds that
\begin{align*}
    \prob[\abs{\ltUsers}=M] &= 
    \prob[\abs{\{\user: \depth\leq l\}}=M]\\
    &= \prob\left[\left(\sum_{\user=1}^{\users} \mathbf{1}_{\depth\leq l}\right)
   =M\right],
\end{align*}
where $\mathbf{1}_{\{\cdot\}}$ is the indicator function. By \ref{itm:random_stragglers}, it follows that $\{ \mathbf{1}_{\depth\leq l}\}_{u=1}^U$ are i.i.d. Bernoulli random variables with parameter $\prob[\mathbf{1}_{\depth\leq l}=1]=\frac{l}{L+1}$, proving \eqref{eq:binom_rv} by the definition of the binomial distribution \cite[Ch. 4]{edition2002probability}. 
%where $\Ber$ denotes a Bernoulli random variable. The first equality stems from \eqref{eq:ltUsers} and the second does by \ref{itm:random_stragglers}, implying that $\prob[\depth\leq l]=\frac{l}{L+1}$.
\end{IEEEproof}

According to Lemma~\ref{obs:binom_rv}, for a given global iteration, the chance that a particular layer would not be updated at all decreases for either deeper layers or growing number of total users $\users$. Lemma~\ref{obs:binom_rv} is thereby used to establish two auxiliary lemmas: The first identifies that the \ac{salf} update rule, which yields a random vector due to the stochastic nature of the stragglers set, is an unbiased estimate of the full \ac{fa} rule, as stated next:

\begin{lemma}[Unbiasedness]\label{lemma:unbiasedness}
When \ref{itm:random_stragglers} holds, for every \ac{fl} round of index $t$ and a given set of training data samples $\{\sample\}$, the global model aggregated via \ac{salf} \eqref{eq:salf_update_rule} with \eqref{eq:p_constant} is an unbiased estimator of the one obtained via vanilla \ac{fa} \eqref{eq:FedAvg_update}, namely,
    \begin{equation}\label{eq:unbiasedness}
        \E[\Tilde\vw_{t+1}]=\vw_{t+1}
    \end{equation}    
\end{lemma}
\begin{IEEEproof}
    The proof is given in Appendix \ref{app:unbiasedness}. 
\end{IEEEproof}
\smallskip

The second auxiliary lemma which follows from Lemma~\ref{obs:binom_rv} bounds the variance of the \ac{salf} \ac{dnn} parameters.

\begin{lemma}[Bounded variance]\label{lemma:bounded_var}
    Consider \ac{salf} where \ref{itm:random_stragglers} holds and $p_l$ is set via \eqref{eq:p_constant}, and for every \ac{fl} round $t$ and a given set of training data samples $\{\sample\}$, the learning rate $\eta_t$ is set to be non-increasing and satisfying $\eta_t \leq 2\eta_{t+1}$. Then, the expected difference between $\Tilde\vw_{t+1}$ and $\vw_{t+1}$, which is  the variance of $\Tilde\vw_{t+1}$, is bounded by
        \begin{equation}\label{eq:bounded_var}
        \E[\|\Tilde\vw_{t+1}-\vw_{t+1}\|^2] \leq \eta_t^2 \frac{4\users LG^2}{(\users-1)}  
    \frac{1+{\left(1-\frac{1}{L+1}\right)}^\users}{1-{\left(1-\frac{1}{L+1}\right)}^\users}.
        \end{equation}
\end{lemma}
\begin{IEEEproof}
    The proof is given in Appendix \ref{app:bounded_var}. 
\end{IEEEproof}
\smallskip

\subsubsection{Convergence Bound}
The characterization of the \ac{dnn} parameters produced by \ac{salf} as a bounded-variance unbiased stochastic estimate of the parameters produced with full \ac{fa}, allows to characterize the convergence of the learning algorithm. 
Using the above notations, the following theorem establishes the convergence bound of \ac{salf}.
\begin{theorem}\label{thm:FL_Convergence}
Consider \ac{salf}-aided \ac{fl} with \eqref{eq:p_constant} satisfying 
\ref{itm:smooth_strng_convx_lip_cont}-\ref{itm:random_stragglers};  define $\kappa=\frac{\rho_s}{\rho_c}, \gamma=\max\{8\kappa,1\},$ and set the learning rate to $\eta_t=\frac{2}{\rho_c(\gamma+t)}$. Then, it holds that
    \begin{align}\label{eq:FL_Convergence}
    &\E\left[F(\Tilde\vw_t)\right] - F(\wOpt)\leq 
    \frac{\kappa}{\gamma+t-1} \times \notag\\
    &\quad\left(\frac{2(B+C)}{\rho_c}  
    +\frac{\rho_c\gamma}{2}\E\left[\|\vw_1-\wOpt\|^2\right]\right),
    \end{align}
    where 
    \begin{subequations}
    \label{eq:C_constant}
    \begin{align}
        B &= \sum^{\users}_{\user=1} \frac{1}{\users^2}\sigma_u^2 +6\rho_s\Gamma;\\
        C &= \frac{4\users L G^2}{(\users-1)}  
    \frac{1+{\left(1-\frac{1}{L+1}\right)}^\users}{1-{\left(1-\frac{1}{L+1}\right)}^\users}.
    \end{align}
    \end{subequations}
\end{theorem}

\begin{IEEEproof}
The proof is given in  Appendix~\ref{app:FL_Convergence}. 
\end{IEEEproof}
\smallskip

Theorem~\ref{thm:FL_Convergence} rigorously bounds the difference between the objective value of a model learned by \ac{salf} at round $t$ to the optimal model $\wOpt$. By setting the step size $\eta_t$ to decrease gradually, which is also known to contribute to the convergence of \ac{fl} \cite{stich2018local, li2019convergence}, Theorem~\ref{thm:FL_Convergence} indicates that \ac{salf} converges at a rate of $\cO(1/t)$. This asymptotic rate is identical to \ac{fl} schemes with no latency limitations \cite{stich2018local, li2019convergence}, stressing the ability of \ac{salf} to carry out low latency \ac{fl} in the presence of dynamic system heterogeneity while mitigating its harmful effects with hardly affecting the learning procedure compared to conventional \ac{fl}. 
%Additionally, as the bound is derived for a general random setting of the stragglers sets, it provides convergence guarantees even if the portion of stragglers is considerable and/or in the presence of large inconsistencies of participants in consecutive layers.  

Nonetheless, in the non-asymptotic regime, the integration of a latency deadline influences model convergence. This is revealed in \eqref{eq:FL_Convergence} by the constant $C$ in~\eqref{eq:C_constant}, that depends on (growing at least linearly in) the number of layers $L$, resulting from \ac{salf}'s layers-wise aggregation technique. This implies that synchronous \ac{fl} with deadline $T_{\max}$, whose values results in some of the users being occasionally stragglers, is expected to converge slower for deeper architectures; for which $C$ is larger compared with shallower \acp{dnn}. 

Such behaviour is also experimentally demonstrated in Section~\ref{sec:experiments}, stemming from the fact that for  \acp{dnn} with many layers, the first few layers will be trained by a small portion of the participating users when operating under a fixed deadline compared with shallow layer. This is in line with \ac{dnn} typical behaviour, as deeper model architectures aim to learn more complex mappings, often require  lengthy learning  and are slower to converge~\cite{glorot2010understanding,zagoruyko2016wide}. 
It is emphasized though that under conventional (non layer-wise) synchronous \ac{fl}, training deep \acp{dnn} under similar low latency constraints is expected to often be infeasible, as effectively all users become stragglers.

\subsection{Discussion}\label{subsec:discussion} 
% Advantages
\ac{salf} is particularly designed for \ac{fl} systems that are constrained to operate with low latency while learning over networks comprised of heterogeneous edge devices. It allows synchronous \ac{fl} operation with small deadline $T_{\max}$, which typically results in a large number of stragglers. This is achieved without notably affecting the conventional \ac{fa}-based \ac{fl} flow by exploiting the recursive nature of the backpropagation to leverage partial gradients for updating the global model. Due to its simple layer-wise aggregation, \ac{salf} asymptotically converges at the same rate as unconstrained \ac{fa}, for a random set of stragglers, while supporting extremely low-latency \ac{fl}. For instance, we numerically show in Section~\ref{sec:experiments} that \ac{salf} can learn accurate models in settings where as much as $90\%$ of the users are stragglers that cannot finish computing their gradients in time.

Our convergence analysis of  \ac{salf} assumes that the local computations of each user behave randomly. The distribution imposed in \ref{itm:random_stragglers} accounts for the fact that a device can become a straggler in a given round not only due to its fundamental hardware, but also due to its availability on that particular round, as edge devices may be occupied also for other tasks than \ac{fl}. Conversely, if we were to utilize \ac{salf}'s layer-wise aggregation using {\em deterministic} sets of stragglers in which the same users update the same layers in each training round, an inevitable bias is expected to arise as some of the overall data would not affect all the layers. In that sense, the stochastic nature of system heterogeneity is leveraged as a contributing factor, eliminating the probability of such a scenario to occur.

% Extensions
The outline of \ac{salf} is based on a generic formulation of synchronous gradient-based \ac{fl}. It only requires the learned model to be a \ac{dnn} trained using backpropagation. While this form of gradient-based learning by far dominates \ac{dnn} training methods to-date, alternative ones such as Kalman-based learning~\cite{chang2023low} and zero-order optimization~\cite{liu2020primer} were also proposed in the literature, for which \ac{salf} would require a dedicated adaptation. 
Additionally, \ac{salf} operates without increasing the complexity at the clients and/or the server compared to conventional \ac{fa}. Hence, its methodology can be combined with other schemes for decreasing \ac{fl} latency via model update compression~\cite{shlezinger2020uveqfed, lang2023compressed, lang2023joint}. 
While our design considers a single local iteration, being tailored to tight deadlines, it can be extended to multiple iterations and possibly combined with proximal-aided aggregation to account for user-varying iterations, e.g., \cite{li2020heterogenous}, on a layer-wise basis. These extensions of \ac{salf} are left for future study.

\section{Experimental Study}\label{sec:experiments}
In this section we numerically evaluate \ac{salf}, and compare it to existing approaches \cite{mcmahan2017communication, bonawitz2019towards, diao2020heterofl, pfeiffer2023federated} tackling system heterogeneity in low-latency \ac{fl}\footnote{The source code used in our experimental study, including all the hyper-parameters, is available online at \url{https://github.com/langnatalie/SALF}.}. Our aim is to experimentally validate that the layer-wise approach of \ac{salf} allows to learn reliable \ac{dnn} models for various architectures in a synchronous low-latency manner. We focus on the training of \acp{dnn} for image processing tasks, where we first consider a simple handwritten digit recognition task (Subsection~\ref{ssec:MNIST}), which allows us to evaluate \ac{salf} in different terms of performance as convergence, accuracy, and latency; in controlled settings. Then, we proceed to a more challenging image classification task (Subsection~\ref{ssec:CIFAR}), where we evaluate performance as well as suitability for different deep architectures. 

% setup
%FL training 
\subsection{Handwritten Digit Recognition}\label{ssec:MNIST}
% data
\subsubsection{Setup}
We first consider the federated training of a handwritten digit classification model using the MNIST dataset \cite{deng2012mnist}. 
The data, comprised of $28 \times 28$ gray-scale images divided into $60,000$  training examples and $10,000$ test examples, is uniformly distributed among $\users=30$ users. 

% arcitectures
{\bf Architectures}:  We train two different \ac{dnn} architectures:
\begin{itemize}
    \item A \ac{mlp} with two hidden layers, intermediate ReLU activations and a softmax output layer.
    \item A \ac{cnn} composed of two convolutional layers and two fully-connected ones, with intermediate ReLU activations, max-pooling layers, and a softmax output layer.
\end{itemize}

{\bf \ac{fl} Training}: 
In each \ac{fl} iteration, the users train the \ac{mlp}/\ac{cnn} model using local mini-batch \ac{sgd} \eqref{eq:FedAvg_update} with learning rate $0.05/0.1$; and the global model is learned using $250/ 150$ \ac{fl} iterations, respectively. To simulate controllable latency-constrained system heterogeneity for different deadlines in a hardware-invariant manner, in each iteration we randomly set a predefined ratio of the users to be stragglers, where for each straggler the depth reached in backpropagation is randomized uniformly (in line with  the  distribution assumed in~\ref{itm:random_stragglers}). 

{\bf \ac{fl} Algorithms}: 
We evaluate the following \ac{fl} methods:
\begin{itemize}
    \item  {\em Vanilla \ac{fl}}, which implements full \ac{fa} without  latency constraints~\cite{mcmahan2017communication} and thus without any stragglers. This approach constitutes the desired performance for the  methods that operate in the presence of stragglers.
    \item {\em Drop-stragglers} \ac{fa}, that discards stragglers~\cite{bonawitz2019towards}.
    \item {\em HetroFL} \cite{diao2020heterofl} which addresses heterogeneous clients by equipping them with corresponding heterogeneous local models with varying computational complexities. To guarantee fair comparison with \ac{salf} in the sense of average number of gradient computations, we set the straggling users to shrink their local models with ratio $0.5$. Comparison to HetroFL is considered only for the \ac{cnn}, as this architecture was covered in \cite{diao2020heterofl}. 
    \item {\em AsyncFL} \cite{pfeiffer2023federated} which has the stragglers participate in the \ac{fa} once they finish the local training rather than at each round. Here, to achieve comparable computations to synchronous settings, a straggler that finished the update up to the $l$th layer within $T_{\max}$, is set to asynchronously participate in the \ac{fa} every $2l$ global iterations.
    \item Our proposed {\em \ac{salf}} (Algorithm~\ref{alg:salf}).  
\end{itemize}

\begin{figure}
\centering    \includegraphics[width=0.9\columnwidth]{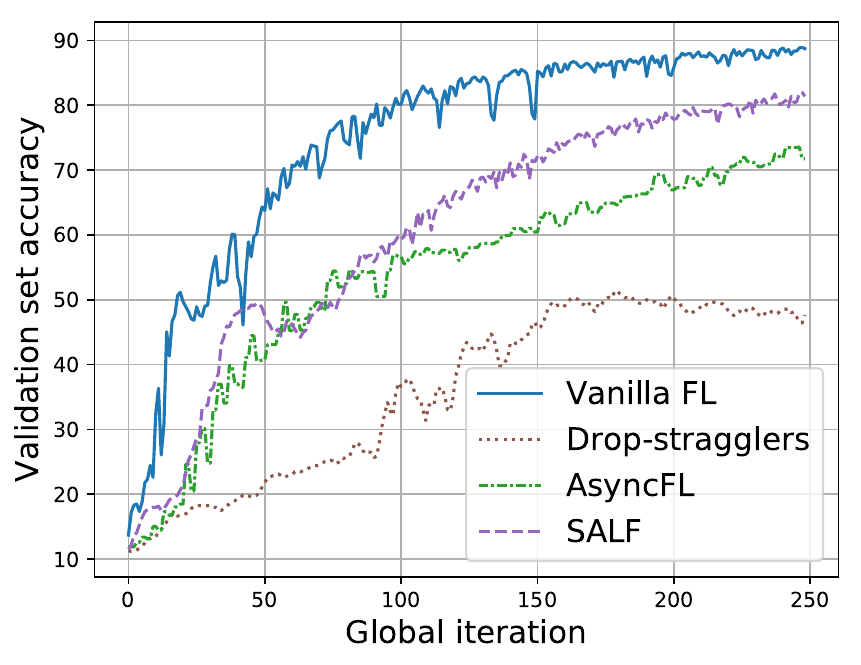}
    \caption{\ac{fl} convergence profile,  \ac{mlp} trained on MNIST.}    \label{fig:convergence_mnist_mlp}   
\end{figure}

\begin{figure}    
\centering    \includegraphics[width=0.9\columnwidth]{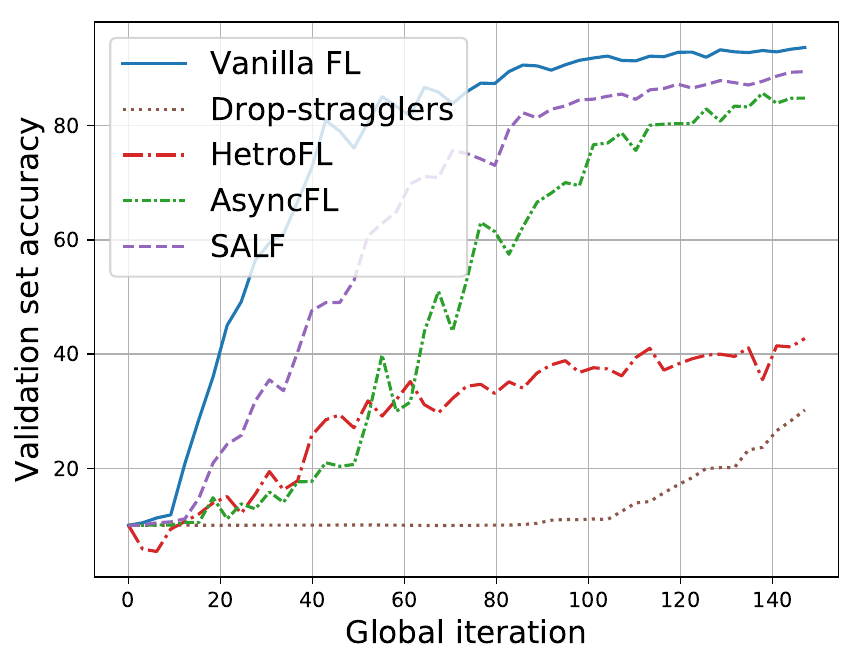}
\caption{\ac{fl} convergence profile,  \ac{cnn} trained on MNIST}    \label{fig:convergence_mnist_cnn2} \end{figure}

\subsubsection{Results}
We compare the performance of the above \ac{fl} methods for the considered  task in their convergence profile (namely, during training); the accuracy of the trained model (i.e., after training is concluded);  and their overall latency.

% summerizing results
%convergence
{\bf Convergence Results}: 
We begin by evaluating the convergence profiles of the learning methods, i.e., the resulting validation accuracy achieved over the training procedure.
To emphasize the gains of \ac{salf} compared with conventional drop-stragglers in facilitating federated training under tight latency constraints, we focus on the extreme case in which $90\%$ of the users are stragglers.
This exemplifies a tight deadline-based \ac{fl} application, where the major bulk of the users effectively become stragglers that do not complete the whole model update. 
Fig.~\ref{fig:convergence_mnist_mlp}-\ref{fig:convergence_mnist_cnn2} illustrate the convergence profile of the methods for the \ac{mlp} and \ac{cnn} models, respectively. There, it is systematically demonstrated that \ac{salf} is the closet to vanilla \ac{fl}, whereas AsyncFL is the second best (while requiring an asynchronous operation). For comparison, drop-stragglers and HetroFL show an inferior performance, as a result of ignoring stragglers contributions for the former, and having the majority of local architectures differ from the archietcture of the global model for the latter. 

{\renewcommand{\arraystretch}{1.2}
\begin{table*}
\caption{Test accuracy results for different stragglers' percents, MNIST dataset.}
\centering
%\setlength{\tabcolsep}{4pt}
%\begin{adjustbox}{width=\columnwidth} 
\begin{tabular}{|c|c|c|c|c|c|c|c|c|c|c|c|c|c|c|c|c|c|}
\hline
%& \multicolumn{9}{c|}{MNIST}\\
%\cline{2-10}
& Vanilla FL & \multicolumn{4}{c|}{Drop-stragglers} & \multicolumn{4}{c|}{\ac{salf}}& \multicolumn{4}{c|}{HetroFL} &
\multicolumn{4}{c|}{AsyncFL}  \\
\cline{2-18}
Straggler $\%$
& N/A 
& 0.3 & 0.5 & 0.7 & 0.9 
& 0.3 & 0.5 & 0.7 & 0.9 
& 0.3 & 0.5 & 0.7 & 0.9
& 0.3 & 0.5 & 0.7 & 0.9 \\
\cline{2-18}
MLP & 0.9
& 0.87 & 0.84 & 0.77 & 0.49
& {\bf 0.88} & {\bf 0.85} & {\bf 0.85} & {\bf 0.81}
& N/A &  N/A &  N/A & N/A 
&  0.86 & 0.84  & 0.82  & 0.73 \\
CNN & 0.95 
& 0.93 & 0.9 & 0.83 & 0.28
& {\bf 0.94} & {\bf 0.93} & {\bf 0.92} & {\bf 0.90} 
& 0.88 &  0.84 &  0.61 & 0.43
& {\bf 0.94} & 0.92   & 0.91  & 0.86 \\
\cline{2-18}
% & \multicolumn{9}{c|}{CIFAR-10}\\
% \cline{2-10}
% CNN2 & 0.44 
% & 0.39 & 0.40 & 0.38 & 0.38
% & 0.43 & 0.43 & 0.43 & 0.42\\ 
\hline
\end{tabular}
%\end{adjustbox}
\label{tbl:summerizing_architectures}
\end{table*}}

{\bf Accuracy Results}: 
We proceed with examining the performance of the trained models in terms of their test accuracy. 
Table~\ref{tbl:summerizing_architectures} summarizes the test accuracy result for both the \ac{mlp} and \ac{cnn} models trained on the MNIST datasets for either $30/50/70/90$ percent of the users being stragglers.  
Table~\ref{tbl:summerizing_architectures} reveals that for all the considered techniques, as expected, the higher the percentage of stragglers, the lower the test accuracy is. This monotonic behaviour results with dramatic degradation for growing percentages (corresponding to tight deadlines) for the synchronous drop-stragglers and HetroFL. Despite that, the degradation of \ac{salf} is notably lower compared to all other baselines, maintaining a minor gap from vanilla \ac{fl} (which operates without latency constraint), and consistently achieving the best performance of the trained model among all stragglers-constrained methods.

    \begin{figure}
     \includegraphics[width=0.9\columnwidth]{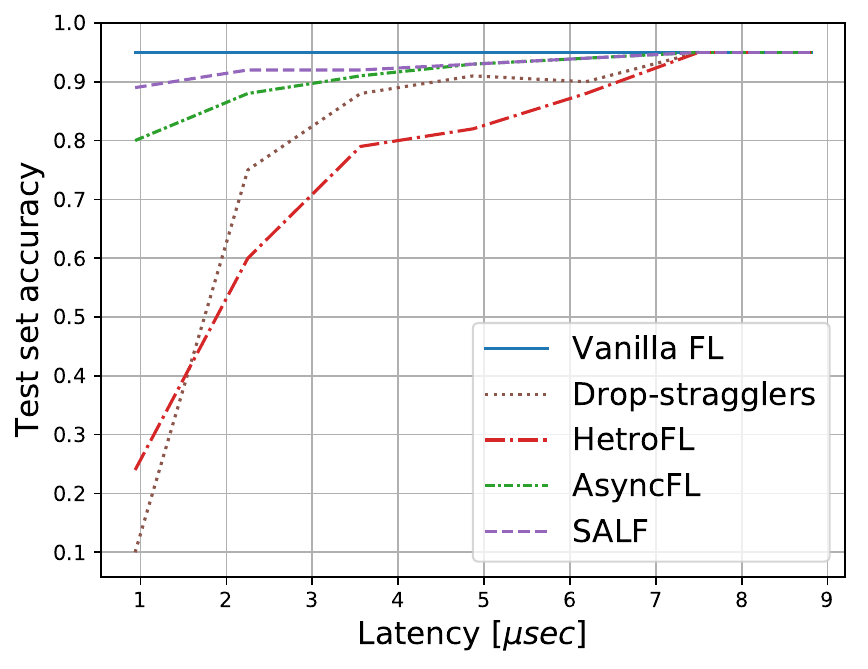}
    \caption{Test set accuracy vs. latency constrains, \ac{cnn} trained on MNIST. 
    }
    \label{fig:latency}
\end{figure}

%latency
{\bf Latency Results}: 
As discussed in Subsection~\ref{subsec:system_heterogeneity_model}, for timing-based \ac{fl} deployments, the deadline $T_{\max}$  determines the latency. 
To translate the improved resilancy to stragglers of \ac{fl} reported above into concrete timings, we evaluate the overall test performance of all considered \ac{fl} methods under different timing constraints  $T_{\max}$. 
We focus on the 2-layer \ac{cnn} model, for which the full backward pass for a single user was empirically evaluated as taking the maximal value of $7.5$ microseconds ($\mu$sec), when computed using a Quadro RTX 6000 GPU.  The resulting test accuracy versus $T_{\max}$ in the range of $[1,9]$$\mu$sec are reported in 
Fig.~\ref{fig:latency} for the considered \ac{fl} methods, in comparison with vanilla \ac{fl} (which operates without latency constraints and is thus invariant of $T_{\max}$). 
%reports the baselines test set accuracy achieved for different latency constrains during the training phase. In particular, lower latency implies lower amount of intermediate layers being updated in the backward pass, where $7.5\mu$ seconds is the estimated time required for a full backward pass to be completed using a Quadro RTX 6000 GPU. As each user performs at most one local iteration, for latency values higher than $7.5\mu$ seconds, the graphs saturate since the backpropagation procedure is done within this time.
%To simulate such a scenario, we linearly mapped lower latency into higher stragglers percents.

We first clearly observe in Fig.~\ref{fig:latency} that, as expected, all \ac{fl} methods coincide with full \ac{fa} when the deadline surpasses the maximal local computation latency of $7.5$ $\mu$sec. This follows since in such latency regimes, none of the users are stragglers. However, in the more interesting regimes of $T_{\max} < 7.5$ $\mu$sec,  we observe that \ac{salf} yields a minor desegregation in performance which hardly grows when the latency is decreased; and outperforms the (asynchronous) AsyncFL counterpart. 
This is in contrast to either drop-stragglers, which is left with hardly any devices updating the global model under tight latency constraints, or HetroFL, where in this case tighter latency implies that the majority of the users train a different model than the global one. 
Specifically, for low $T_{\max}$, i.e., below $1.5$ $\mu$sec, which also corresponds to high percentages of straggling clients, HetroFL is superior to drop-stragglers in performance (where the opposite holds for moderate values of $T_{\max}$, i.e., for $1.5< T_{\max}<7$ $\mu$sec), 
and AsyncFL is superior to both; aligned with similar findings in Fig.~\ref{fig:convergence_mnist_cnn2} and Table~\ref{tbl:summerizing_architectures}.

Finally, as evidenced in Fig.~\ref{fig:latency}, for the tightest latency value, \ac{salf} realizes a drop of merely $5\%$ from the accuracy of vanilla FL, compared to $15\%,70\%$ and 
$90\%$ in the case of the AsyncFL, HetroFL, and drop-stragglers, respectively.
Consequently, the gains of \ac{salf} in performance are persistent, and most dominant in the low-latency regime, where stragglers mostly fail to meet the deadline calculation time $T_{\max}$ and their partial updates are harnessed for modifying the global model by the layer-wise approach of \ac{salf}.

\subsection{Image Classification}\label{ssec:CIFAR}
\subsubsection{Setup}
We proceed to evaluating \ac{salf} in tasks typically requiring deeper \acp{dnn} compared to the ones used in the previous subsection. Here, \ac{fl} is implemented for the distributed training of natural image classification model using the CIFAR-10 dataset \cite{CIFAR-10}.
This set is comprised of $32 \times 32$ RGB images divided into $50,000$ training examples and $10,000$ test examples.

{\bf Architecture}:
We explore whether the learning profile of a model trained via the layer-wise aggregation of \ac{salf} changes with varying the depth of its given architecture. To that aim, we use the VGG model architecture \cite{simonyan2014very} with four depths, namely, number of convolutional layers, that are $11$, $13$, $16$, and $19$.  \ac{fl} training is similar to the one described in Subsection~\ref{ssec:MNIST}, while using a learning rate of $0.05$ and set the amount of global iterations to be $1,500$ for each architecture.

\begin{figure*}
    \centering    
    \includegraphics[width=0.45\textwidth]{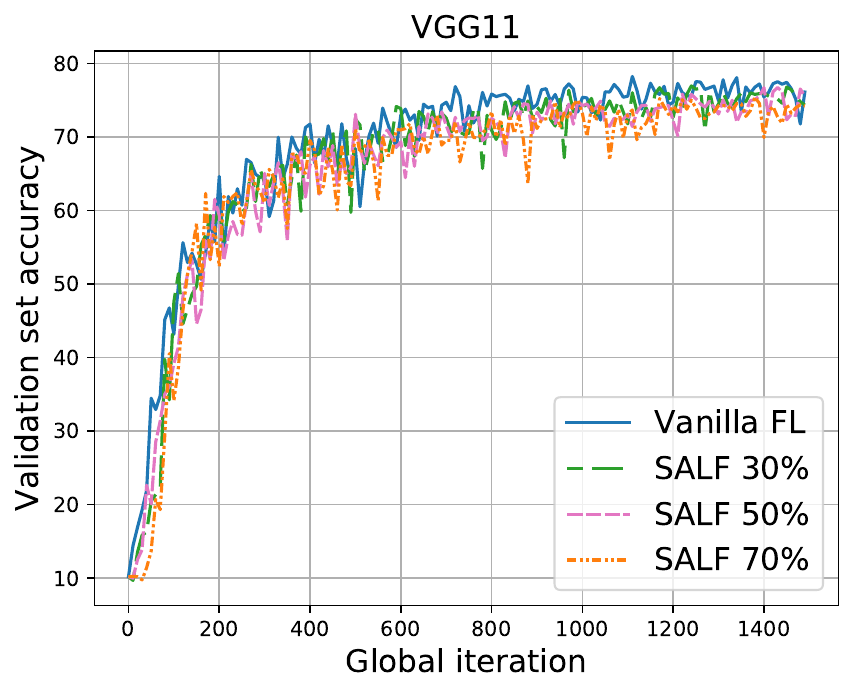}
    \includegraphics[width=0.45\textwidth]{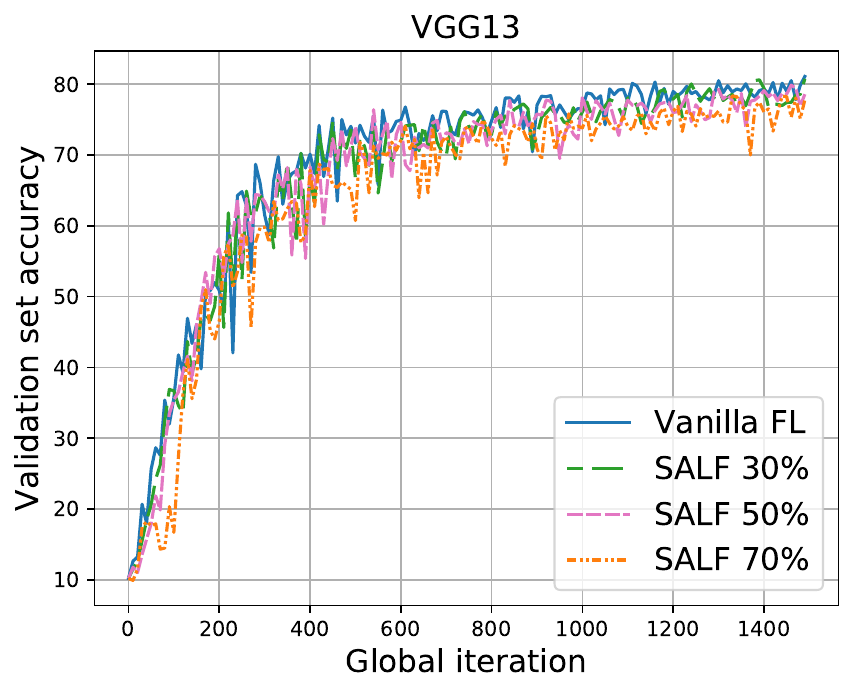}
    \includegraphics[width=0.45\textwidth]{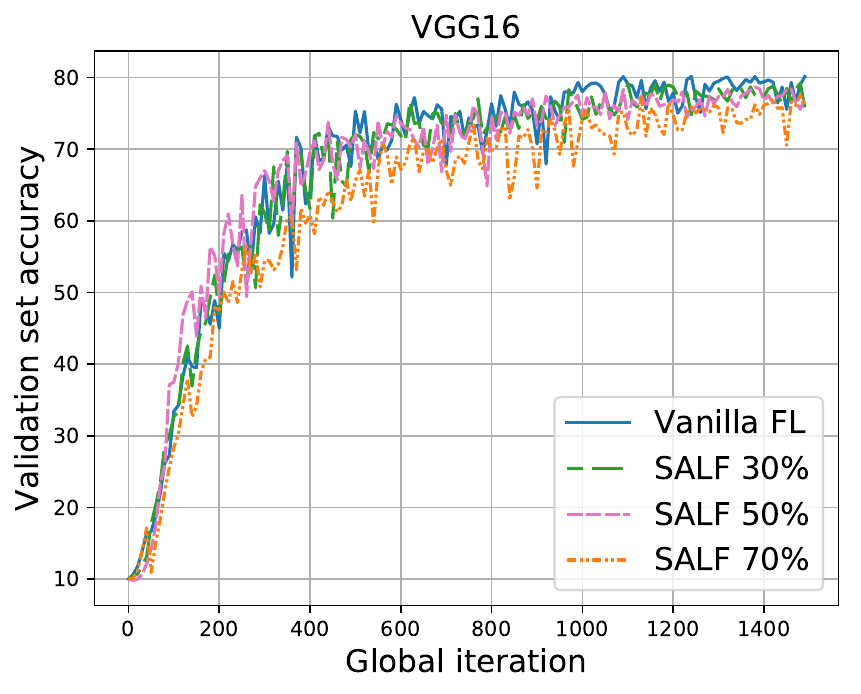}
    \includegraphics[width=0.45\textwidth]{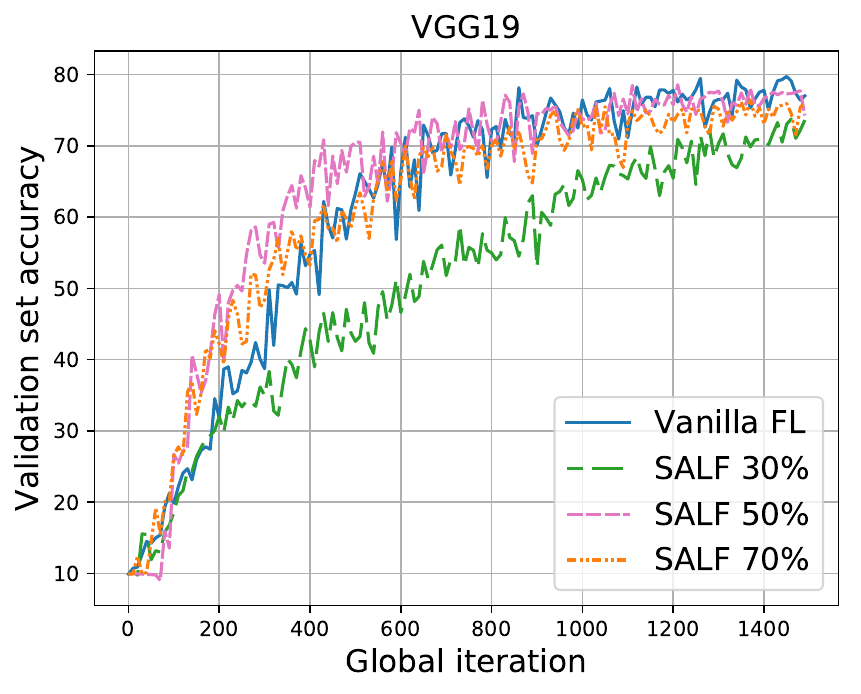} 
       \caption{Convergence profile of \ac{fl} schemes training VGG models using the CIFAR-10 dataset for different stragglers percentages.}
   \label{fig:vggPrecentage} 
\end{figure*}

\subsubsection{Results}
%{\bf Convergence Results}: 
We first evaluate \ac{salf} for different timing constraints. Fig.~\ref{fig:vggPrecentage} shows the convergence profile of the VGG models for vanilla FL and \ac{salf} operating with either $30/50/70$ percents of straggling clients. It can be generally observed that, \ac{salf} converges for all considered architectures and amount of stragglers. However, as discussed in Subsection~\ref{subsec:discussion}, the training of deeper architectures exhibit different profiles compared with shallower ones, which are consistent for all considered stragglers percentages. A notable phenomenon is observed when training the deepest model of VGG19; There, \ac{salf} with $70\%$ or $50\%$ stragglers yields an improved learning procedure compared with $30\%$. This can be associated with the fact that when training deep models, adding minor levels of distortion, which in our case result from the growth in the variance of the stochastic estimate in Lemma~\ref{lemma:bounded_var} for deeper networks, can lead to improving the converged model, in accordance with similar findings in \cite{Guozhong1995NoiseBackprop,sery2021over}. 

\begin{figure*}
    \centering    
    \includegraphics[width=0.45\textwidth]{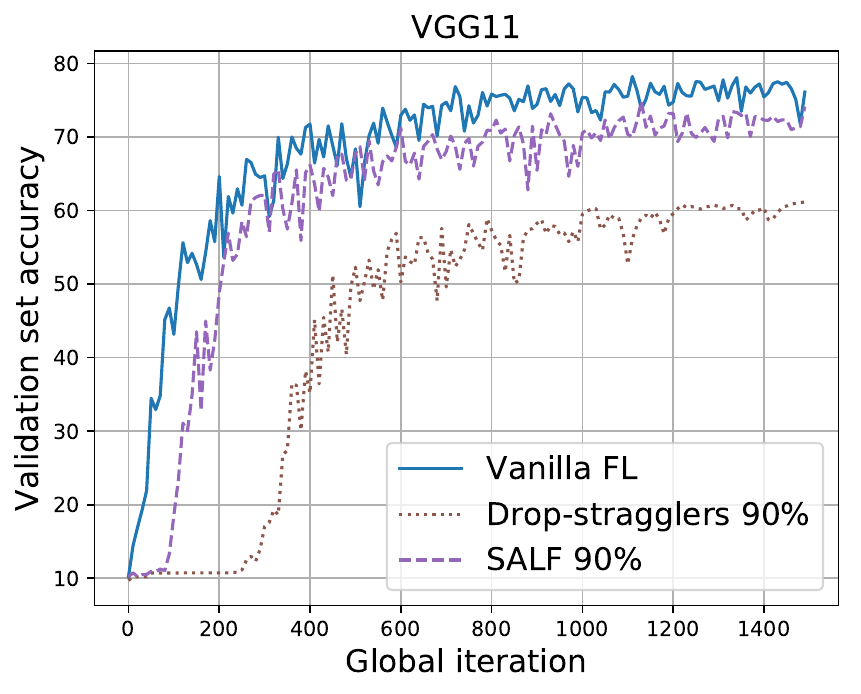}
    \includegraphics[width=0.45\textwidth]{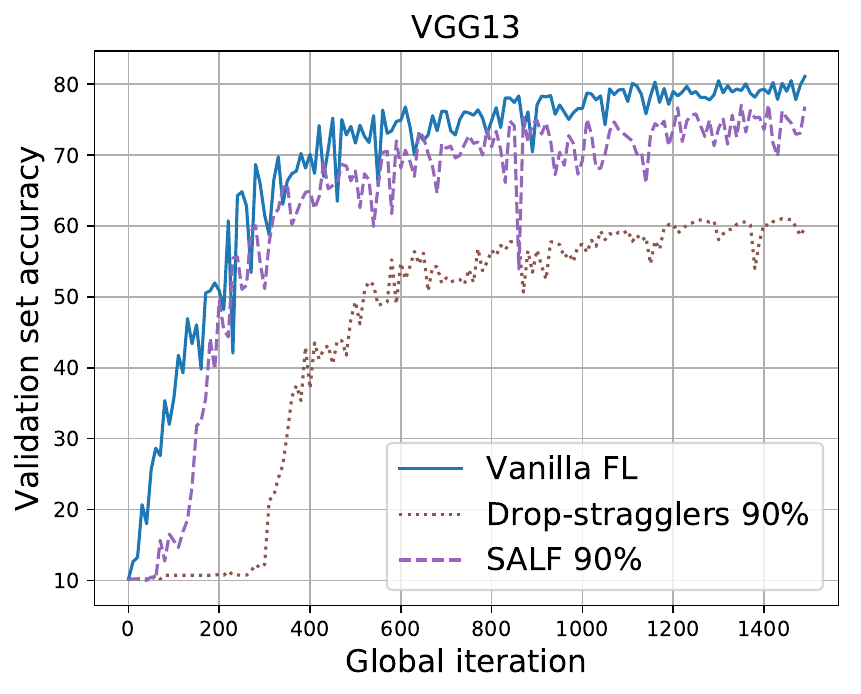}\\
    \includegraphics[width=0.45\textwidth]{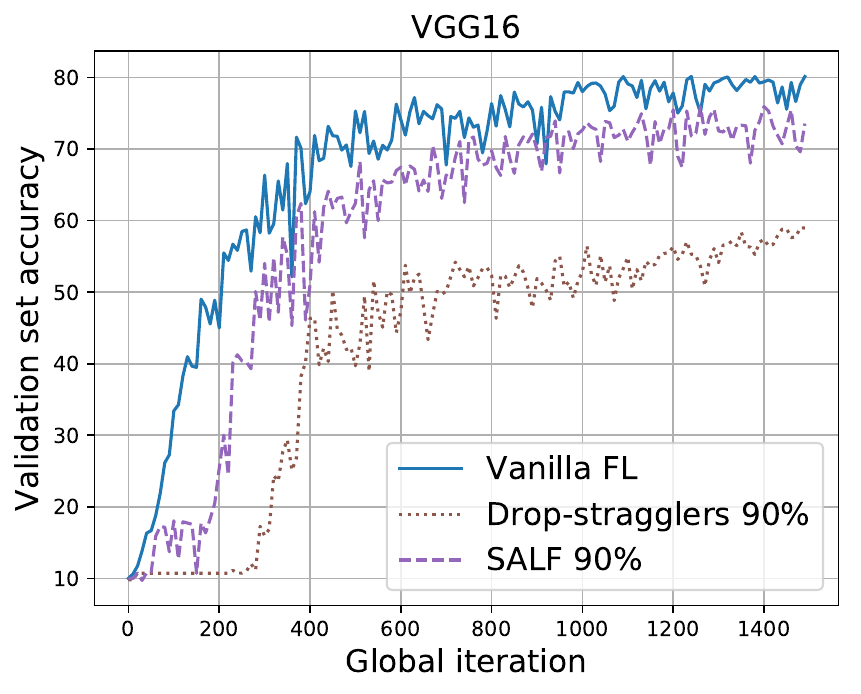}
    \includegraphics[width=0.45\textwidth]{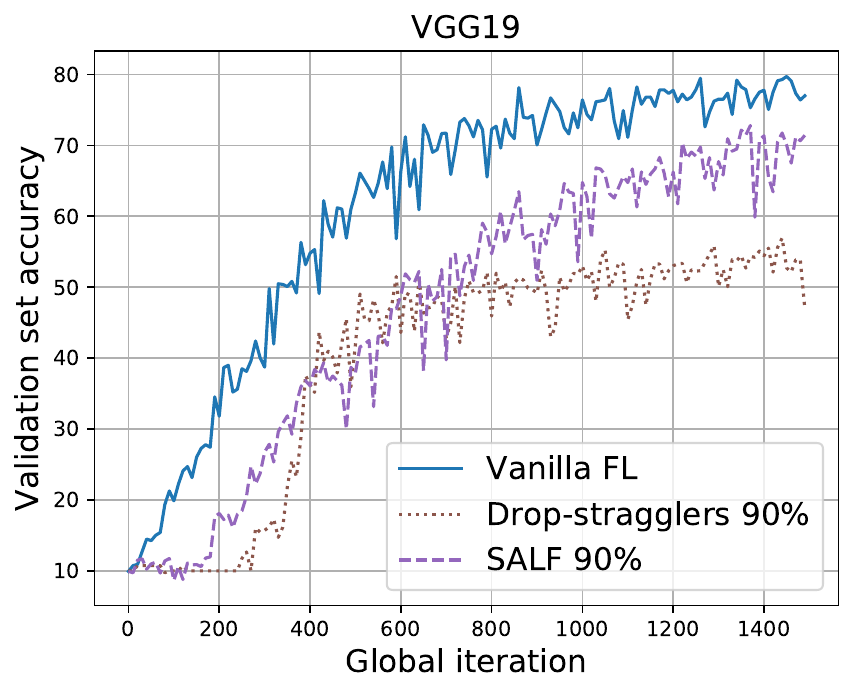}    
   \caption{Convergence profile of \ac{fl} schemes training VGG models using the CIFAR-10 dataset at $90\%$ stragglers percentage.}
   \label{fig:vgg90}
\end{figure*}

To better highlight the interplay between our layer-wise \ac{fl} and the overall \ac{dnn} depth, we conclude our study by focusing merely on a challenging setup with tight deadlines. 
In Fig.~\ref{fig:vgg90}, we depict \ac{salf} and drop-stragglers for $90\%$ straggles. The results in Fig.~\ref{fig:vgg90}  stress the power of \ac{salf} in mitigating their harmful effect significantly better than drop-stragglers, also for various deep architectures, similarly to the findings evidenced in Subsection~\ref{ssec:MNIST}. In addition, the convergence profiles of Fig.~\ref{fig:vgg90}, once observed in comparison with~Fig.~\ref{fig:convergence_mnist_mlp}-\ref{fig:convergence_mnist_cnn2}, numerically support Theorem~\ref{thm:FL_Convergence} under the non-asymptotic regime; indeed indicating that low-latency \ac{fl} converges slower with deeper \ac{dnn}s, while systematically exceeding drop-stragglers and still approaching the performance achieved with stragglers-free full \ac{fa}.

\section{Conclusions}\label{sec:conclusions}
In this work we proposed \ac{salf}, which is an \ac{fl} algorithm that implements layer-wise global aggregation to incorporate stochastically determined straggling users in tight timings synchronous settings. \ac{salf} utilizes the last-to-first layer update policy of backpropagation-based \ac{dnn} training to exploit partial gradients and update each of the model layers separately, with possibly different  amount of users in each. We analyzed the convergence profile of \ac{salf} accompanied by numerical evaluations, demonstrating that it operates reliably under tight latency constraints and approaches the performance achieved by \ac{fl} with no stragglers.

\section*{Acknowledgments}
The authors are grateful to Dor Elimelech for fruitful discussions and helpful ideas regarding the probabilistic modelling. 

\begin{appendix}
\numberwithin{lemma}{subsection} 
\numberwithin{corollary}{subsection} 
\numberwithin{remark}{subsection} 
\numberwithin{equation}{subsection}	

\subsection{Proof of Lemma~\ref{lemma:unbiasedness}}\label{app:unbiasedness} 
In order to show that $\Tilde\vw_{t+1}$ \eqref{eq:salf_update_rule} is an unbiased estimator of $\vw_{t+1}$ \eqref{eq:FedAvg_update}, it suffices to show that this holds for the $l$th sub-vector of both, according to the decomposition of a vector into its $L$ sub-vectors in \eqref{eq:global_layers}. 
Assuming that the random data sample indexes $\{\sample\}$ are given, the only source of randomness in $\vw^l_{t+1}$ is encapsulated in $\ltUsers$. Yet, instead of calculating the expectation of $\Tilde\vw^l_{t+1}$ with respect to $\ltUsers$, we leverage the observation that $\abs{\ltUsers}$ is a binomial random variable \eqref{eq:binom_rv}, and utilize the law of total expectation, yielding 
\begin{align}\label{eq:unbiaesedness_total_expectation2}   
\E[\Tilde\vw^l_{t+1}]
&=\E\left[\E\left[\Tilde\vw^l_{t+1} \middle| \abs{\ltUsers}\right]\right] \notag\\
&=\sum_{K=0}^\users 
 \prob\left[\abs{\ltUsers} = K\right] \cdot \E\left[\Tilde\vw^l_{t+1} \middle| \abs{\ltUsers} = K\right]\notag\\
&= \prob\left[\abs{\ltUsers} = 0\right] \cdot \E\left[\Tilde\vw^l_{t+1} \middle| \abs{\ltUsers} = 0\right] \notag\\
 &+\sum_{K=1}^\users 
 \prob\left[\abs{\ltUsers} = K\right] \cdot \E\left[\Tilde\vw^l_t \middle| \abs{\ltUsers} = K\right].
\end{align}
Next, we note that for the distribution of $\abs{\ltUsers}$ in Lemma~\ref{obs:binom_rv}, the definition of $p_l$ in~\eqref{eq:p_constant} holds that 
\begin{equation}
\label{eqn:p0}
    p_l=\prob\left[\abs{\ltUsers} = 0\right].
\end{equation}
Substituting \eqref{eqn:p0} combined with \ac{salf}'s aggregation rule in \eqref{eq:salf_update_rule} into \eqref{eq:unbiaesedness_total_expectation2} results in 
\begin{align}\label{eq:unbiaesedness_total_expectation}   
\E[\Tilde\vw^l_{t+1}]
&= p_l \cdot \vw^l_t
 + \sum_{K=1}^\users 
 \prob\left[\abs{\ltUsers} = K\right] \frac{1}{1-p_l} \times \notag\\
 &\left(\frac{1}{K}
\E\left[\sum_{\user\in \ltUsers} \vw^l_{\user,t}  \middle| \abs{\ltUsers} = K\right] -p_l\vw^l_t\right).
\end{align}

Notice that, according to \ref{itm:random_stragglers}, the distribution of $\ltUsers$ given that $\abs{\ltUsers}=K$ is uniform over all size-$K$-out-of-size-$\users$ subsets. 
In~\cite[Lem. 4]{li2019convergence}, the authors proved 
that \ac{fa} with partial device participation, such that a fixed-size amount of users are sampled uniformly in a without replacement fashion, is an unbiased estimator of the full device participation alternative. Equivalently,   
\begin{align}\label{eq:unbiased_given_K}
    \frac{1}{K}
\E\left[\sum_{\user\in \ltUsers} \vw^l_{\user,t}  \middle| \abs{\ltUsers} = K\right] &= \frac{1}{K}
K\sum_{\user=1}^\users\frac{1}{\users}\vw^l_{\user,t} = \vw^l_{t+1}. 
\end{align} 
Plugging  \eqref{eq:unbiased_given_K} into \eqref{eq:unbiaesedness_total_expectation} and the fact that 
\begin{equation*}
 \sum_{K=1}^\users 
 \prob\left[\abs{\ltUsers}= K\right]  = 1- \prob\left[\abs{\ltUsers} = 0\right] = 1-p_l,   
\end{equation*}
results with 
\begin{align*}
    \E[\Tilde\vw^l_{t+1}] &=  p_l \cdot \vw^l_t + \left(\vw^l_{t+1} -  p_l \cdot \vw^l_t\right) \frac{1}{1-p_l} (1-p_l)   \\
    &=\vw^l_{t+1},
\end{align*}
thus proving \eqref{eq:unbiasedness}.
 \qed

\subsection{Proof of Lemma~\ref{lemma:bounded_var}}\label{app:bounded_var}
By the definition of the $\ell_2$ norm, it follows that the variance of $\Tilde\vw_{t+1}$ is the sum over all its $1,\dots,L$ sub vectors (layers) variances; and, similarly to Appendix~\ref{app:unbiasedness}, we have that
\begin{align}\label{eq:layer_vise_variance}
    &\mathbb{E}\left[\|\Tilde\vw^l_{t+1}-\vw^l_{t+1}\|^2\right]=
    \E\left[\E\left[\|\Tilde\vw^l_{t+1}-\vw^l_{t+1}\|^2 \middle| \abs{\ltUsers}\right]\right]\notag\\
    &=\prob\left[\abs{\ltUsers} = 0\right] \E\left[\|\Tilde\vw^l_{t+1}-\vw^l_{t+1}\|^2 \middle| \abs{\ltUsers} = 0\right] 
    + \notag\\
    &\sum_{K=1}^\users \prob\left[\abs{\ltUsers} = K\right] \cdot
     \E\left[\|\Tilde\vw^l_{t+1}-\vw^l_{t+1}\|^2 \middle| \abs{\ltUsers} = K\right]. 
\end{align}
For the first summoned, it holds that
\begin{align}\label{eq:first_summond_bound}
&\E\left[\|\Tilde\vw^l_{t+1}-\vw^l_{t+1}\|^2 \middle| \abs{\ltUsers} = 0\right] =\notag\\
&\E\left[\left\|\vw^l_t-\vw^l_{t+1}\right\|^2 \right] 
\overset{(a)}{\leq}
\sum_{\user=1}^\users \frac{1}{\users} \E\left[\left\|
\left(\vw^l_{\user,t} -\vw^l_t\right)\right\|^2 \right] =  \notag\\
&\sum_{\user=1}^\users \frac{1}{\users} \E\left[\left\|
\eta_t\nabla F_\user(\vw_t,\sample)\right\|^2 \right] \overset{(b)}{=} 
\eta^2_tG^2,
\end{align}
where $(a)$ follows by the convexity of $\|\cdot\|^2$ and $(b)$ stems from the bounded norm assumed by \ref{itm:bounded_norm}.

As for the second summoned \eqref{eq:layer_vise_variance}, it can be written as
\begin{align}
\E&\left[\left\|\Tilde\vw^l_{t+1}-\vw^l_{t+1}\right\|^2 \middle| \abs{\ltUsers} = K\right] = \notag\\
\E&\left[\left\|
\frac{1}{1-p_l}\left(\sum_{\user\in \ltUsers} \frac{1}{\abs{\ltUsers}}\
\vw^l_{\user,t} 
-p_l\vw^l_t\right)
-\vw^l_{t+1}\right\|^2 \middle| \abs{\ltUsers} = K\right] \notag\\  
&=\frac{1}{(1-p_l)^2} \times \notag\\
\E&\left[\left\|
    \sum_{\user\in \ltUsers} \frac{1}{\abs{\ltUsers}}
    \vw^l_{\user,t} -\vw^l_{t+1}
    -p_l(\vw^l_t -\vw^l_{t+1})
    \right\|^2 \middle| \abs{\ltUsers} = K\right] \notag\\ 
    &=\frac{1}{(1-p_l)^2} \times \notag\\
    \Biggl(
    &\E\left[\left\|
    \sum_{\user\in \ltUsers} \frac{1}{\abs{\ltUsers}}
    \vw^l_{\user,t} -\vw^l_{t+1}    
    \right\|^2 \middle| \abs{\ltUsers} = K\right]  + \label{eq:term_I} \\
    &p_l^2\E\left[\left\|
    \vw^l_t -\vw^l_{t+1}
    \right\|^2 \middle| \abs{\ltUsers} = K\right] + \label{eq:term_II} \\
    &-p_l(\vw^l_t -\vw^l_{t+1})^T
    \E\left[
    \sum_{\user\in \ltUsers} \frac{1}{\abs{\ltUsers}}
    \vw^l_{\user,t} -\vw^l_{t+1}    
     \middle| \abs{\ltUsers} = K\right]
    \Biggr).\label{eq:cross_term}
\end{align}
Now, $\eqref{eq:cross_term}=0$ by \eqref{eq:unbiased_given_K}; $\eqref{eq:term_II}\leq p_l^2\eta_t^2G^2$ by \ref{itm:bounded_norm}; and finally, \eqref{eq:term_I} can be bounded using the result obtained in ~\cite[Lem. 5]{li2019convergence} due to the same reasons mentioned in Appendix~\ref{app:unbiasedness}. That is, by \cite[Lem. 5]{li2019convergence} it holds that
\begin{align*}
    \E\left[\left\|
    \sum_{\user\in \ltUsers} \frac{1}{\abs{\ltUsers}}
    \vw^l_{\user,t} -\vw^l_{t+1}    
    \right\|^2 \middle| \abs{\ltUsers} = K\right]  
    \leq \\
\frac{\users}{K(\users-1)}\left(1-\frac{K}{\users}\right)4\eta_t^2G^2.
\end{align*}

Overall, \eqref{eq:layer_vise_variance} is thus given by
\begin{align}\label{eq:second_summond_bound}
    &\eqref{eq:layer_vise_variance} \leq \frac{1}{(1-p_l)^2} \sum_{K=1}^\users \prob\left[\abs{\ltUsers} = K\right]  \times \notag\\        
    &\left(\frac{\users}{K(\users-1)}\left(1-\frac{K}{\users}\right)4\eta_t^2G^2 + p_l^2\eta^2_tG^2 \right) \notag\\
    &\leq     
    \frac{1}{(1-p_l)^2} \sum_{K=1}^\users 
    \prob\left[\abs{\ltUsers} = K\right] \cdot 
    \left(
    \frac{\users}{(\users-1)}4\eta_t^2G^2 
    + p_l^2\eta^2_tG^2
    \right) \notag\\
    &= \eta_t^2 \frac{G^2\left(\frac{4\users}{(\users-1)}
    + p_l^2\right)}{1-p_l}.
\end{align}
Adding both bounds of \eqref{eq:first_summond_bound} and \eqref{eq:second_summond_bound} while summing over all $L$  layers results with
\begin{align*}
\mathbb{E}[\|\Tilde\vw_{t+1}-\vw_{t+1}\|^2] 
=\sum_{l=1}^L\mathbb{E}\left[\|\Tilde\vw^l_{t+1}-\vw^l_{t+1}\|^2\right]\\
\leq \sum_{l=1}^L 
\eta_t^2 \frac{G^2\left(\frac{4\users}{(\users-1)}
    + p_l\right)}{1-p_l} \leq 
\eta_t^2 G^2\frac{4\users}{(\users-1)}
    \sum_{l=1}^L  \frac{1+p_l}{1-p_l} \\ \leq
    \eta_t^2 G^2\frac{4\users}{(\users-1)}L  
    \frac{1+{\left(1-\frac{1}{L+1}\right)}^\users}{1-{\left(1-\frac{1}{L+1}\right)}^\users},
\end{align*}
where the last inequality follows by bounding each fraction in the summation with the one obtained from setting the largest numerator and the smallest denominator; proving \eqref{eq:bounded_var}.
\qed

\subsection{Proof of Theorem~\ref{thm:FL_Convergence}}\label{app:FL_Convergence}
By modelling the model updates of \ac{salf} as unbiased stochastic estimates of \ac{fa} (Lemma~\ref{lemma:unbiasedness}) with bounded variance (Lemma~\ref{lemma:bounded_var}), we recast our setting of \ac{fl} with random layer-wise computations as \ac{fl} with random partial participation as considered in~\cite[Thm. 3]{li2019convergence}. Accordingly, by replacing Lemmas $4, 5$ of \cite{li2019convergence} with our Lemmas~\ref{lemma:unbiasedness}, \ref{lemma:bounded_var}, respectively, we obtain that the derivation of~\cite[Thm. 3]{li2019convergence} applies for our setting of \ac{salf} with the corresponding coefficients in \eqref{eq:C_constant}, thus proving the theorem.
\qed
\end{appendix}

\bibliographystyle{IEEEtran}
\bibliography{IEEEabrv,refs}

\end{document}